\documentclass[runningheads]{llncs}

\usepackage[T1]{fontenc}
\usepackage[utf8]{inputenc}
\usepackage{amsmath,amssymb}
\usepackage{graphicx}
\usepackage{tikz}
\usetikzlibrary{shapes.geometric, arrows.meta, positioning, fit, backgrounds, calc}
\usepackage{booktabs}
\usepackage{multirow}
\usepackage{caption}
\usepackage{subcaption}
\usepackage{hyperref}
\usepackage{cite}
\usepackage{color}
\usepackage{balance}
\usepackage{comment}
\usepackage{marvosym} 

\begin{document}

\title{UAVCAN-Graph: Temporal Graph Construction and Residual GNN
       for Multi-Class Attack Detection in UAV Systems}

\titlerunning{UAVCAN-Graph: Residual GNN for Attack Detection in UAV Systems}

\author{Md Farhan Shahriyar\inst{1}$^{*}$ \and
        Muhammad Naeem Tabassam\inst{2} \and
        Shujahat Anwar\inst{3} \and
        Muhammad Shahroz Khalid\inst{3} \and
        S M Shahriar Rahman\inst{4}}

\authorrunning{M. F. Shahriyar et al.}

\institute{
  Hajee Mohammad Danesh Science and Technology University,
  Dinajpur, Bangladesh\\
  \email{farhanshahriyar@gmail.com}
  \and
  College of Physics and Optoelectronics Engineering,\\
  Shenzhen University, Shenzhen 518060, China\\
  \email{naeemtabassam@gmail.com}
  \and
  Shenzhen University, Shenzhen 518060, China\\
  \email{shujahatanwar55@gmail.com, 2351471004@email.szu.edu.cn}
  \and
  Brac University, Dhaka, Bangladesh\\
  \email{shahriar168529@gmail.com}\\[4pt]
  $^{*}$Corresponding author
}

\maketitle

\begin{abstract}
Unmanned Aerial Vehicles (UAVs) rely on Controller Area Network (CAN) bus
communication governed by the UAVCAN protocol for real-time data exchange
between onboard components. This communication channel remains vulnerable
to cyber attacks including flooding, fuzzy injection and replay attacks.
Existing intrusion detection approaches predominantly employ binary
classification that fails to distinguish among specific attack types. In
this paper we propose UAVCAN-GRAPH, a graph-based intrusion detection
framework that transforms temporal CAN bus traffic into communication graphs
using a sliding window approach. Each graph encodes node-level features
derived from inter-arrival time statistics, payload entropy, UAVCAN transfer
ID continuity and PageRank centrality. A Residual GraphSAGE architecture
with skip connections processes these graphs through four message-passing
layers followed by dual pooling for graph-level classification into four
classes: Normal, Flooding, Fuzzy and Replay. We employ Focal Loss to handle
class imbalance and augment underrepresented attack categories through dense
temporal windowing. Evaluation on the HCRL UAVCAN Attack Dataset comprising
10 attack scenarios demonstrates that UAVCAN-GRAPH achieves a macro-averaged
ROC-AUC of 0.9954 with strong separation across all four attack categories.
The t-SNE visualization of learned graph embeddings confirms distinct
clustering of attack types in the latent space.

\end{abstract}

\section{Introduction}

Unmanned Aerial Vehicles have experienced rapid adoption across commercial, military and civilian domains including surveillance, precision agriculture and package delivery \cite{shakhatreh2019unmanned}. Modern UAV architectures depend on the Controller Area Network bus as the primary communication backbone connecting flight controllers, sensors, actuators and GPS modules \cite{koubaa2019micro}. The UAVCAN protocol operates at the application layer over CAN bus to provide standardized message formatting for real-time avionics data exchange \cite{kim2022uavcan}.

The broadcast nature of CAN bus communication creates inherent security vulnerabilities. Any node connected to the bus can transmit messages without authentication and there is no built-in mechanism to verify message origin or integrity \cite{wu2019survey}. Attackers exploiting these weaknesses can inject malicious packets through flooding attacks that overwhelm the bus with high-frequency messages, fuzzy attacks that inject packets with randomized payloads or replay attacks that retransmit previously captured legitimate messages \cite{majumder2024controller}.

Traditional intrusion detection systems for CAN bus networks rely on statistical thresholds or machine learning classifiers that operate on tabular feature representations of individual packets or fixed-length sequences \cite{song2020vehicle}. These approaches treat network traffic as independent samples and fail to capture the relational structure inherent in multi-node communication patterns. Furthermore, most existing works formulate the detection problem as binary classification distinguishing only between normal and attack traffic without identifying the specific attack type \cite{avatefipour2019intelligent}.ML-based detection has succeeded across diverse threat domains
\cite{ chy2026email, chy2025clocmlguard}; however, CAN bus IDS
approaches remain limited to flat feature representations that
fail to capture inter-node communication topology.

Graph Neural Networks (GNNs) offer a principled approach to modeling relational data by operating directly on graph-structured inputs \cite{wu2020comprehensive}. Recent studies have demonstrated the potential of GNNs for network intrusion detection by representing traffic flows as graphs where nodes correspond to network entities and edges encode communication relationships \cite{lo2022graphsage}. However, the application of GNNs to CAN bus security and specifically to the UAVCAN protocol remains largely unexplored.

In this paper we propose UAVCAN-GRAPH, a framework that addresses these limitations through three contributions:
\begin{enumerate}
    \item A temporal graph construction method that transforms sliding windows of CAN bus traffic into directed weighted graphs where nodes represent unique CAN IDs and edges capture consecutive message transitions.
    \item A Residual GraphSAGE architecture with skip connections and dual pooling that performs graph-level multi-class classification into Normal, Flooding, Fuzzy and Replay categories.
    \item A training strategy combining Focal Loss with minority-class augmentation through dense temporal windowing to address inherent class imbalance in attack datasets.
\end{enumerate}

The remainder of this paper is organized as follows. Section~II reviews related work. Section~III presents the proposed methodology. Section~IV describes experimental setup and results. Section~V concludes the paper.

\section{Related Work}

Early intrusion detection approaches for CAN bus relied on rule-based methods monitoring message frequency and timing anomalies \cite{wu2019survey}. Song et al. \cite{song2020vehicle} proposed a deep convolutional neural network operating on fixed-length sequences of CAN messages to detect DoS, fuzzy and impersonation attacks. Avatefipour et al. \cite{avatefipour2019intelligent} developed a multi-stage approach combining feature extraction with ensemble classifiers for binary anomaly detection on automotive CAN traffic. These methods process traffic as flat sequential data without modeling the communication topology among bus participants.

Graph Neural Networks have gained attention for network security applications due to their ability to capture structural relationships. Lo et al. \cite{lo2022graphsage} applied GraphSAGE-based models to flow-level network intrusion detection and demonstrated improvements over traditional feature-based classifiers. Chang et al. \cite{chang2021graph} proposed E-GraphSAGE that incorporates edge features into the message passing framework for network traffic anomaly detection. For CAN bus specifically, Majumder et al. \cite{majumder2024controller} introduced a graph construction approach where temporal transitions between CAN IDs form edges and PageRank centrality serves as a node-level feature for binary attack detection on UAVCAN traffic. Their work established the viability of graph-based representations for CAN bus IDS but was limited to binary classification without distinguishing attack types.

The HCRL UAVCAN Attack Dataset \cite{kim2022uavcan} provides labeled CAN bus traffic from UAVs under flooding, fuzzy and replay attacks across 10 scenarios. Recent work \cite{park2024uavcan} applied reinforcement based   model  to multiclass dataset for attack detection. The framework combined  architectures for UAVCAN anomaly detection in IDS system. However, multi-class classification that distinguishes among specific attack types on this dataset has not been addressed using graph-based approaches.

\section{Proposed Methodology}

\begin{figure*}[!t]
\centering
\resizebox{\textwidth}{!}{
\begin{tikzpicture}[
    node distance=0.6cm and 0.8cm,
    block/.style={rectangle, draw=black!70, fill=blue!8, text width=2.4cm, minimum height=1.1cm, align=center, rounded corners=3pt, font=\small\sffamily},
    bigblock/.style={rectangle, draw=black!70, fill=green!8, text width=2.8cm, minimum height=1.3cm, align=center, rounded corners=3pt, font=\small\sffamily},
    lossblock/.style={rectangle, draw=black!70, fill=red!8, text width=2.4cm, minimum height=1.1cm, align=center, rounded corners=3pt, font=\small\sffamily},
    outblock/.style={rectangle, draw=black!70, fill=orange!12, text width=2.6cm, minimum height=1.1cm, align=center, rounded corners=3pt, font=\small\sffamily},
    datablock/.style={rectangle, draw=black!70, fill=gray!12, text width=2.6cm, minimum height=1.1cm, align=center, rounded corners=3pt, font=\small\sffamily},
    arr/.style={-{Stealth[length=2.5mm]}, thick, black!70},
    skipline/.style={-{Stealth[length=2.5mm]}, thick, black!40, dashed},
    grouplabel/.style={font=\footnotesize\sffamily\bfseries, text=black!60},
]

\node[datablock] (raw) {Raw .bin Files\\(10 Scenarios)};
\node[block, right=of raw] (parse) {UAVCAN\\Parser};
\node[block, right=of parse] (label) {Multi-Class\\Labeling};

\node[block, right=of label] (window) {Sliding Window\\(W=64, S=32)};
\node[bigblock, right=of window] (graph) {Temporal Graph\\Construction};
\node[block, right=of graph] (feat) {Node Feature\\Extraction\\(15 features)};

\node[block, right=of feat] (proj) {Input\\Projection};
\node[bigblock, right=of proj] (sage) {Residual\\GraphSAGE\\(4 layers)};
\node[block, right=of sage] (pool) {Mean + Max\\Pooling\\+ Density};

\node[lossblock, right=of pool] (focal) {Focal Loss\\($\gamma = 2$)};
\node[outblock, right=of focal] (out) {4-Class\\Output\\N / F / Fz / R};

\draw[arr] (raw) -- (parse);
\draw[arr] (parse) -- (label);
\draw[arr] (label) -- (window);
\draw[arr] (window) -- (graph);
\draw[arr] (graph) -- (feat);
\draw[arr] (feat) -- (proj);
\draw[arr] (proj) -- (sage);
\draw[arr] (sage) -- (pool);
\draw[arr] (pool) -- (focal);
\draw[arr] (focal) -- (out);

\draw[skipline] ([yshift=0.55cm]proj.north) -- node[above, font=\tiny\sffamily, text=black!50] {skip connections} ([yshift=0.55cm]sage.north);

\node[grouplabel, above=0.9cm of parse] {Data Pipeline};
\node[grouplabel, above=0.9cm of graph] {Graph Construction};
\node[grouplabel, above=0.9cm of sage] {Residual GNN};
\node[grouplabel, above=0.9cm of out] {Classification};

\end{tikzpicture}
}
\caption{Overview of the UAVCAN-GRAPH methodology. Raw CAN bus traffic is parsed and labeled, then transformed into temporal graphs via sliding windows. The Residual GraphSAGE model with skip connections processes graph-structured data for four-class attack classification using Focal Loss.}
\label{fig:methodology}
\end{figure*}
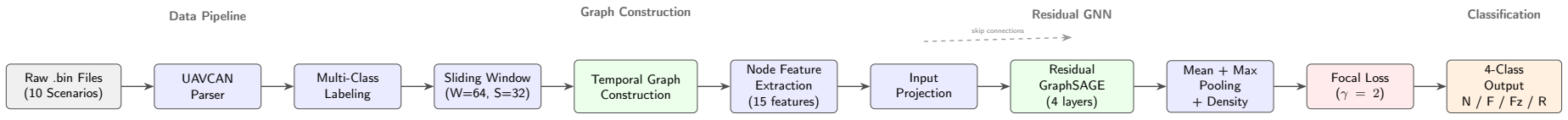

Fig.~\ref{fig:methodology} illustrates the complete UAVCAN-GRAPH pipeline comprising four stages: data preprocessing, temporal graph construction, Residual GraphSAGE classification and multi-class output with Focal Loss optimization.

\subsection{Data Preprocessing and Multi-Class Labeling}

The HCRL UAVCAN Attack Dataset \cite{kim2022uavcan} provides CAN bus traffic captured from UAVs equipped with PX4 autopilot systems across 10 attack scenarios. Each record contains a binary Normal/Attack label, a timestamp, a CAN identifier, the data length code and an 8-byte payload. Scenarios 1--2 contain flooding attacks, scenarios 3--4 contain fuzzy injection attacks, scenarios 5--6 contain replay attacks and scenarios 7--10 contain mixed combinations of multiple attack types.

We assign multi-class labels based on scenario metadata. For pure-attack scenarios (1--6) the binary attack label maps directly to the corresponding attack category. For mixed scenarios (7--10) we define temporal windows that delineate attack phases based on time offsets from the trace start. Attack packets falling within a defined window receive the corresponding label while remaining attack packets outside explicit windows are assigned to the nearest temporal attack phase.

\subsection{Temporal Graph Construction}

We transform sequential CAN bus traffic into graph-structured data using a sliding window approach. Given a window of $W$ consecutive CAN messages with stride $S$, we construct a directed weighted graph $G = (V, E, \mathbf{X})$ where:

\begin{itemize}
    \item $V$ is the set of unique CAN identifiers observed within the window.
    \item $E$ contains directed edges $(v_i, v_j)$ for each pair of consecutively occurring CAN IDs where $v_i$ appears immediately before $v_j$ in the message sequence.
    \item Edge weight $w_{ij}$ counts the number of times the transition $(v_i, v_j)$ occurs within the window.
\end{itemize}

For each node $v \in V$ we extract a 15-dimensional feature vector comprising four groups of features.

\textbf{Temporal features} capture message timing patterns including message count, transmission frequency and inter-arrival time statistics (mean, standard deviation, minimum and maximum). Flooding attacks produce abnormally high frequencies with compressed inter-arrival times.

\textbf{Payload statistics} encode byte-level characteristics including payload Shannon entropy, byte value mean, standard deviation, minimum, maximum and zero-byte ratio. Fuzzy attacks inject random payloads that elevate entropy values while replay attacks maintain consistent payload distributions.

\textbf{UAVCAN protocol features} exploit the tail byte structure of UAVCAN frames. The transfer ID field (lower 5 bits) should increment sequentially modulo 32 under normal operation. We compute transfer ID continuity as the proportion of sequential increments and the toggle bit change ratio from bit 5 of the tail byte. Injected frames disrupt these protocol-specific patterns.

\textbf{Graph structural features} include PageRank centrality computed through 20 power iterations with a damping factor of 0.85. We also compute graph density as the ratio of observed edges to maximum possible edges which serves as a graph-level feature appended during classification.

The graph-level label is determined by majority voting among attack packets within the window. If more than 30\% of packets carry attack labels the most frequent attack class becomes the graph label; otherwise the graph is labeled as Normal.

\subsection{Residual GraphSAGE Architecture}

Standard GNN architectures suffer from over-smoothing when stacked beyond two or three layers as repeated neighborhood aggregation causes node representations to converge \cite{hamilton2017inductive}. We address this limitation through a Residual GraphSAGE architecture that incorporates skip connections to preserve discriminative features across layers.

The input feature matrix $\mathbf{X} \in \mathbb{R}^{|V| \times 15}$ first passes through a linear projection layer:
\begin{equation}
    \mathbf{H}^{(0)} = \text{ReLU}(\mathbf{X}\mathbf{W}_{\text{proj}} + \mathbf{b}_{\text{proj}})
\end{equation}
where $\mathbf{W}_{\text{proj}} \in \mathbb{R}^{15 \times d}$ maps features to a hidden dimension $d$.

Each subsequent layer $l \in \{1, \ldots, L\}$ applies GraphSAGE convolution with a residual connection:
\begin{equation}
    \mathbf{H}^{(l)} = \mathbf{H}^{(l-1)} + \text{Dropout}\Big(\text{ReLU}\big(\text{BN}(\text{SAGE}(\mathbf{H}^{(l-1)}, E))\big)\Big)
\end{equation}
where SAGE denotes the GraphSAGE aggregation operator \cite{hamilton2017inductive} and BN represents batch normalization. The additive residual term ensures gradient flow through deeper architectures.

After $L=4$ message-passing layers we obtain graph-level representations through dual readout pooling:
\begin{equation}
    \mathbf{z}_G = [\text{MeanPool}(\mathbf{H}^{(L)}) \| \text{MaxPool}(\mathbf{H}^{(L)}) \| \delta_G]
\end{equation}
where $\|$ denotes concatenation and $\delta_G$ is the graph density scalar. The concatenated vector passes through a three-layer MLP classifier with a bottleneck architecture ($d \rightarrow d/2 \rightarrow 4$) producing logits for four output classes.

\subsection{Focal Loss and Class Balancing}

Attack datasets exhibit significant class imbalance where normal traffic dominates and certain attack types are underrepresented. Standard cross-entropy loss assigns equal importance to all samples causing the model to favor majority classes. We adopt Focal Loss \cite{lin2017focal} defined as:
\begin{equation}
    \text{FL}(p_t) = -\alpha_t (1 - p_t)^{\gamma} \log(p_t)
\end{equation}
where $p_t$ is the predicted probability for the true class, $\gamma$ is the focusing parameter and $\alpha_t$ provides per-class weighting. Setting $\gamma = 2$ reduces the loss contribution from well-classified samples by a factor of $(1-p_t)^2$ allowing the model to concentrate on difficult boundary cases such as fuzzy attack graphs that share characteristics with other attack types.

We further address imbalance through data augmentation. For underrepresented classes we generate additional training graphs by reducing the sliding window stride from 32 to 16 creating overlapping windows with 75\% overlap that provide varied temporal contexts of the same attack traffic.

\section{Experimental Evaluation}

\subsection{Dataset Description}

We evaluate UAVCAN-GRAPH on the HCRL UAVCAN Attack Dataset \cite{kim2022uavcan} consisting of 10 scenarios captured from drones equipped with PX4 autopilot systems. Table~\ref{tab:scenarios} summarizes the dataset characteristics.

\begin{table}[!t]
\centering
\caption{HCRL UAVCAN Attack Dataset Overview}
\label{tab:scenarios}
\begin{tabular}{@{}ccrrc@{}}
\toprule
Scenario & Attack Type & Total Pkts & Attack Pkts & Ratio \\
\midrule
1 & Flooding & 207,858 & 116,816 & 56.2\% \\
2 & Flooding & 134,170 & 31,930 & 23.8\% \\
3 & Fuzzy & 197,479 & 95,878 & 48.6\% \\
4 & Fuzzy & 133,374 & 29,170 & 21.9\% \\
5 & Replay & 180,608 & 50,612 & 28.0\% \\
6 & Replay & 241,321 & 81,088 & 33.6\% \\
7 & Mixed & 234,162 & 92,612 & 39.6\% \\
8 & Mixed & 265,800 & 115,308 & 43.4\% \\
9 & Mixed & 230,378 & 67,252 & 29.2\% \\
10 & Mixed & 207,380 & 75,850 & 36.6\% \\
\midrule
\multicolumn{2}{c}{\textbf{Total}} & \textbf{2,032,530} & \textbf{756,516} & \textbf{37.2\%} \\
\bottomrule
\end{tabular}
\end{table}

\subsection{Implementation Details}

We set the sliding window size $W = 64$ and stride $S = 32$. Graph construction yields a total of over 50,000 graphs across four classes. We split the data into training (68\%), validation (12\%) and test (20\%) sets using stratified sampling to maintain class proportions.

The Residual GraphSAGE model uses a hidden dimension $d = 192$ with $L = 4$ layers and dropout rate of 0.3. Training proceeds for up to 120 epochs using AdamW optimizer with initial learning rate $8 \times 10^{-4}$, weight decay $5 \times 10^{-4}$ and cosine annealing warm restarts (initial period $T_0 = 20$, multiplier $T_{\text{mult}} = 2$). Early stopping with patience of 20 epochs monitors validation macro-F1. All features are standardized using statistics computed from the training set. Experiments run on a single NVIDIA T4 GPU.

\subsection{Multi-Class Classification Results}

Table~\ref{tab:multiclass} presents the per-class classification performance on the test set. The model achieves strong precision and recall across all four categories with Normal traffic classified most reliably.

\begin{table}[!t]
\centering
\caption{Multi-Class Classification Performance}
\label{tab:multiclass}
\begin{tabular}{@{}lcccc@{}}
\toprule
Class & Precision & Recall & F1-Score & Support \\
\midrule
Normal & 0.9998 & 0.9732 & 0.9863 & 4,995 \\
Flooding & 0.8557 & 0.9828 & 0.9149 & 1,515 \\
Fuzzy & 0.9438 & 0.8708 & 0.9058 & 3,430 \\
Replay & 0.8968 & 0.9542 & 0.9246 & 2,706 \\
\midrule
Macro Avg & 0.9240 & 0.9453 & 0.9329 & 12,646 \\
Weighted Avg & 0.9453 & 0.9425 & 0.9427 & 12,646 \\
\bottomrule
\end{tabular}
\end{table}

Flooding achieves the highest recall at 0.9828 due to its distinctive high-frequency pattern that produces clearly anomalous temporal features. Fuzzy attacks present the greatest classification challenge with an F1-score of 0.9058. This difficulty arises because fuzzy injection randomizes payloads without a fixed signature making it harder to distinguish from other attack types that also alter payload statistics.

\subsection{Binary Detection Performance}

When aggregating the three attack classes into a single Attack category the model achieves binary detection performance shown in Table~\ref{tab:binary}. The near-perfect attack recall of 0.9999 indicates that the model misses almost no attack traffic.

\begin{table}[!t]
\centering
\caption{Binary Detection Performance (Normal vs. Attack)}
\label{tab:binary}
\begin{tabular}{@{}lcccc@{}}
\toprule
Class & Precision & Recall & F1-Score & Support \\
\midrule
Normal & 0.9998 & 0.9732 & 0.9863 & 4,995 \\
Attack & 0.9828 & 0.9999 & 0.9913 & 7,651 \\
\midrule
Macro Avg & 0.9913 & 0.9865 & 0.9888 & 12,646 \\
\bottomrule
\end{tabular}
\end{table}

\subsection{Training Dynamics}

\begin{figure}[!t]
\centering
\includegraphics[width=\columnwidth]{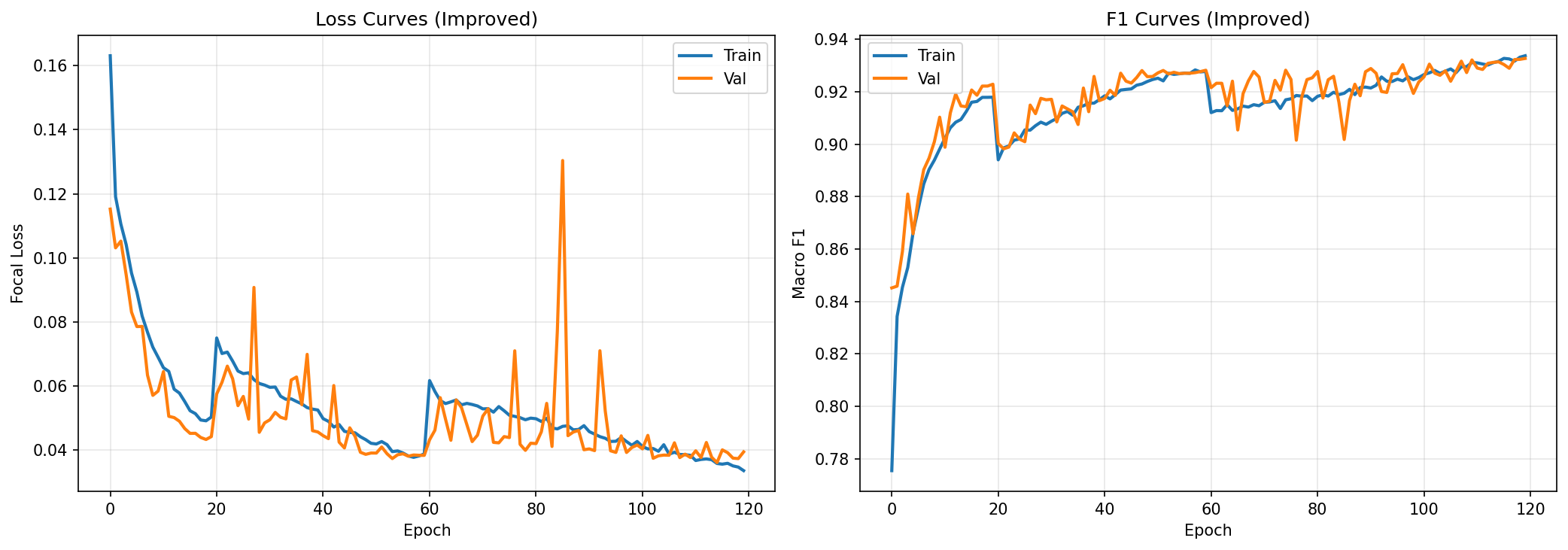}
\caption{Training and validation curves showing Focal Loss (left) and macro-F1 score (right) over 120 training epochs. Learning rate warm restarts are visible as perturbations near epochs 20 and 60.}
\label{fig:training}
\end{figure}

Fig.~\ref{fig:training} shows the Focal Loss and macro-F1 curves over training epochs. The loss decreases steadily with visible perturbations at epochs 20 and 60 corresponding to the cosine annealing warm restart points where the learning rate is reset. These restarts allow the model to escape local minima and explore new regions of the parameter space. The F1 score stabilizes above 0.92 after approximately 40 epochs.

\subsection{ROC and Precision-Recall Analysis}

\begin{figure}[!t]
\centering
\includegraphics[width=\columnwidth]{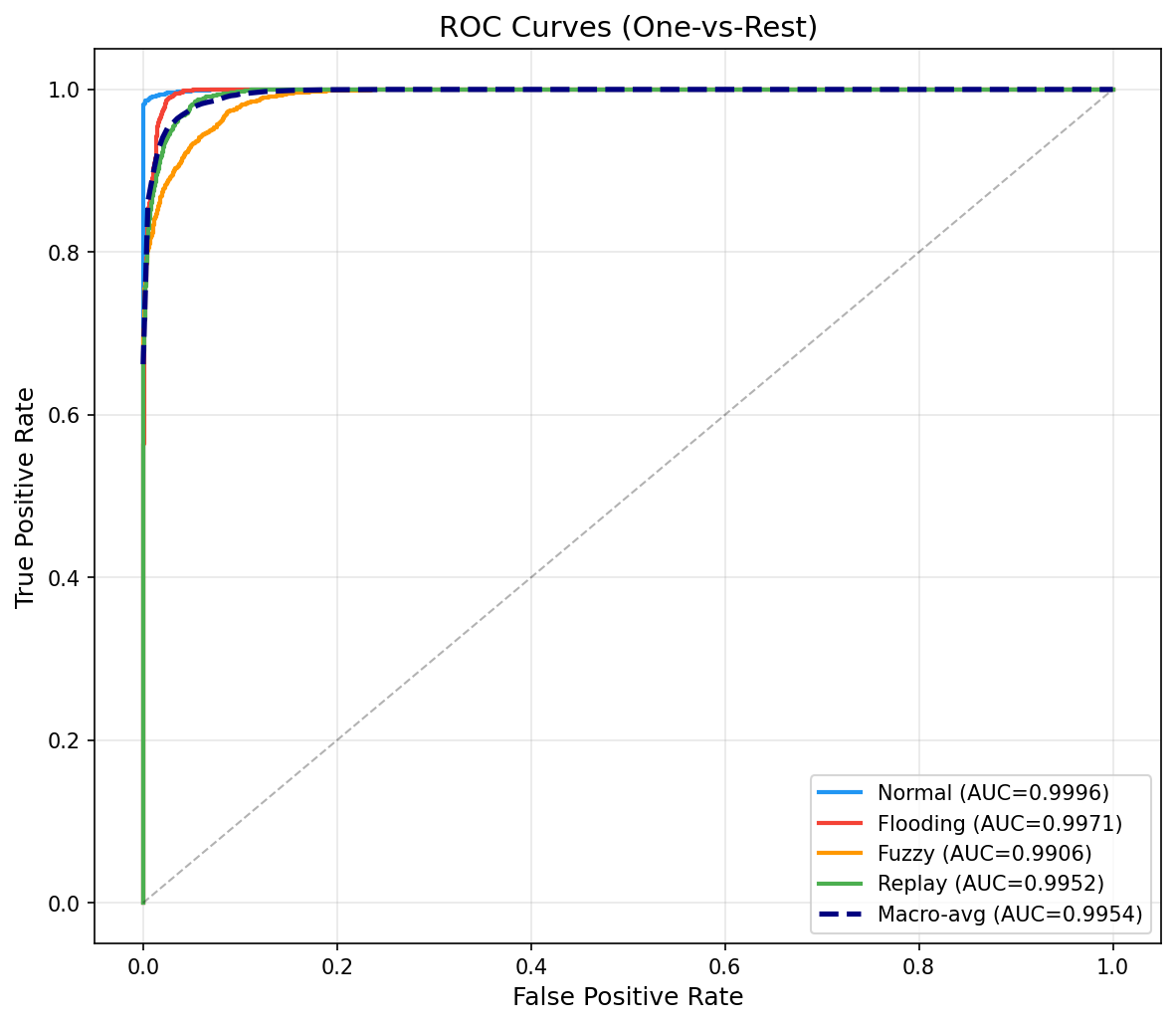}
\caption{One-vs-Rest ROC curves for each attack class and macro-averaged performance.}
\label{fig:roc}
\end{figure}

Fig.~\ref{fig:roc} presents the one-vs-rest ROC curves for each class. All four classes achieve AUC values exceeding 0.99 with Normal traffic reaching the highest AUC. The macro-averaged AUC demonstrates strong discriminative capability across all categories.

\begin{figure}[!t]
\centering
\includegraphics[width=\columnwidth]{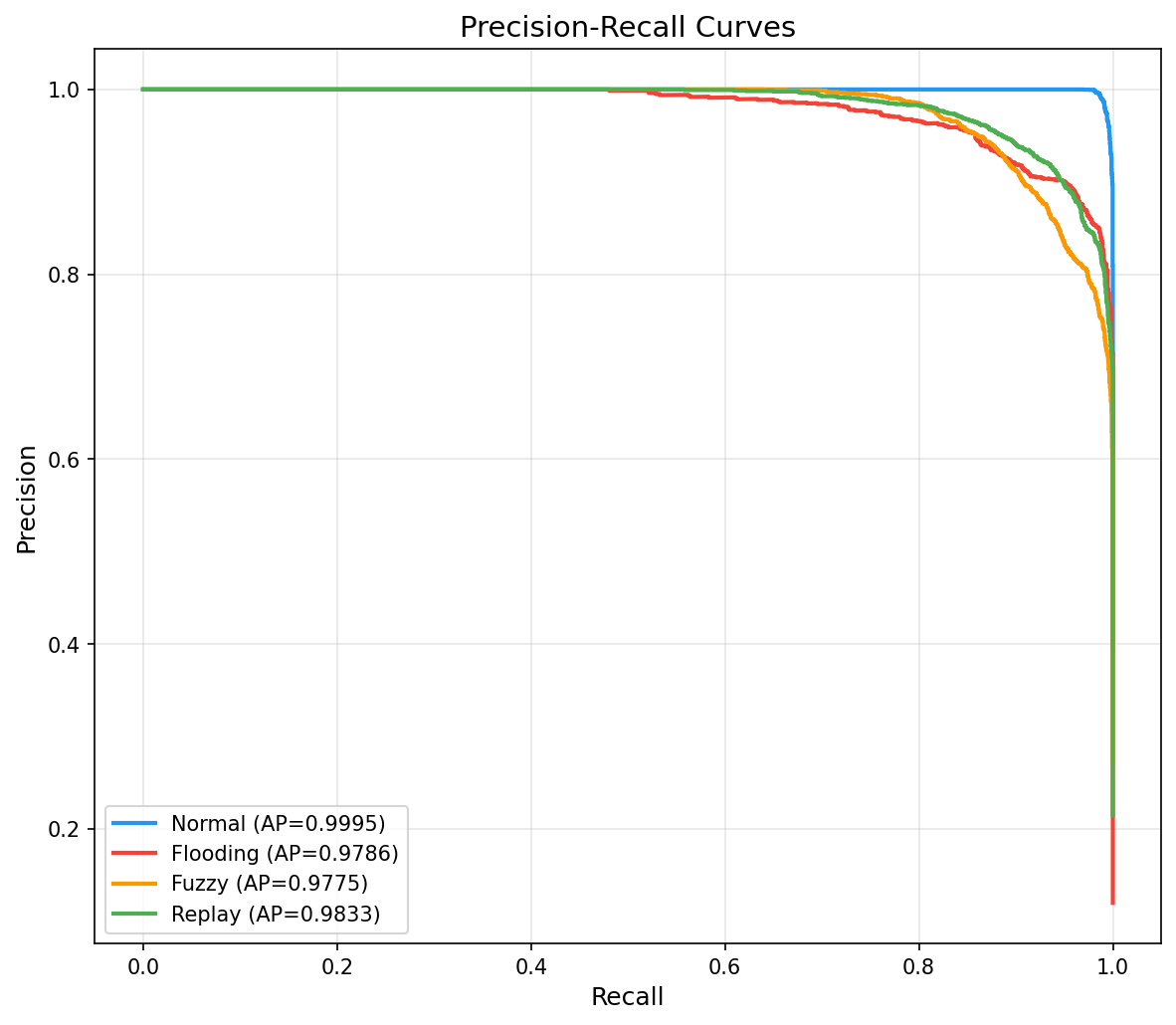}
\caption{Precision-Recall curves showing per-class average precision scores.}
\label{fig:pr}
\end{figure}

Fig.~\ref{fig:pr} shows the precision-recall curves where all classes maintain high average precision. The Fuzzy class exhibits the earliest precision drop at high recall levels reflecting the challenge of classifying every fuzzy attack graph without introducing false positives from neighboring classes.

\subsection{Confusion Matrix Analysis}

\begin{figure}[!t]
\centering
\includegraphics[width=\columnwidth]{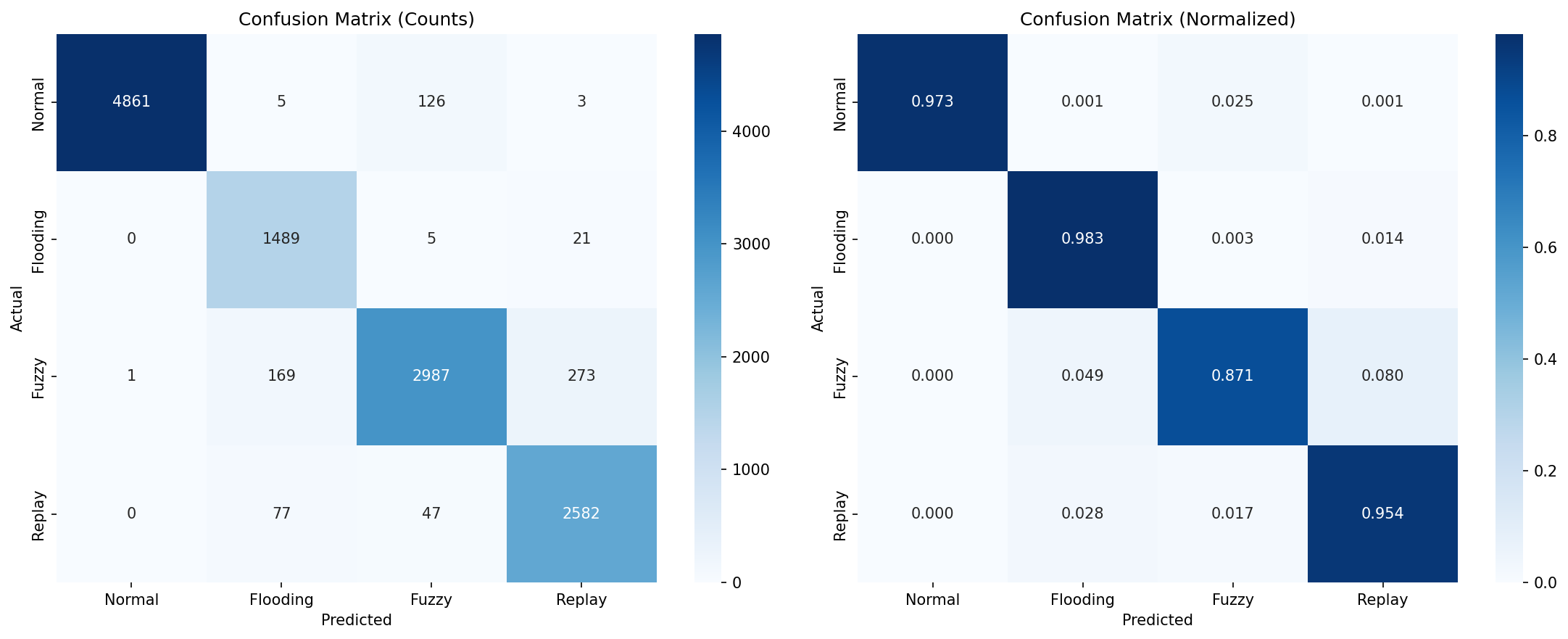}
\caption{Multi-class confusion matrix showing raw counts (left) and row-normalized proportions (right) for all four classes.}
\label{fig:cm}
\end{figure}

Fig.~\ref{fig:cm} shows the multi-class confusion matrix in both raw count and normalized formats. Normal traffic is classified with 97.3\% accuracy. The primary source of confusion occurs between Fuzzy and Replay classes where 273 Fuzzy graphs are misclassified as Replay and 77 Replay graphs are misclassified as Flooding. This inter-attack confusion is expected since mixed scenarios contain transitions between attack types within the same traffic trace.

\begin{figure}[!t]
\centering
\includegraphics[width=0.65\columnwidth]{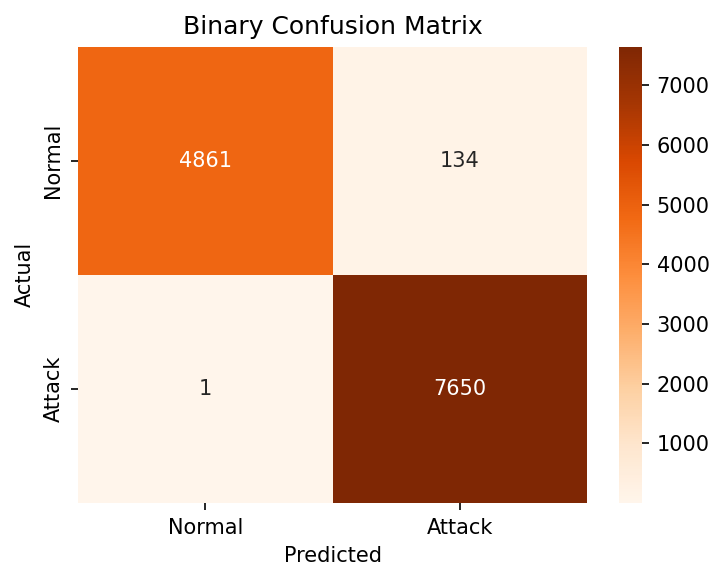}
\caption{Binary confusion matrix for Normal versus Attack classification.}
\label{fig:binary_cm}
\end{figure}

Fig.~\ref{fig:binary_cm} presents the binary confusion matrix confirming that only 1 attack graph is misclassified as Normal out of 7,651 total attack graphs demonstrating the reliability of the system for safety-critical deployment.

\subsection{Embedding Space Visualization}

\begin{figure}[!t]
\centering
\includegraphics[width=\columnwidth]{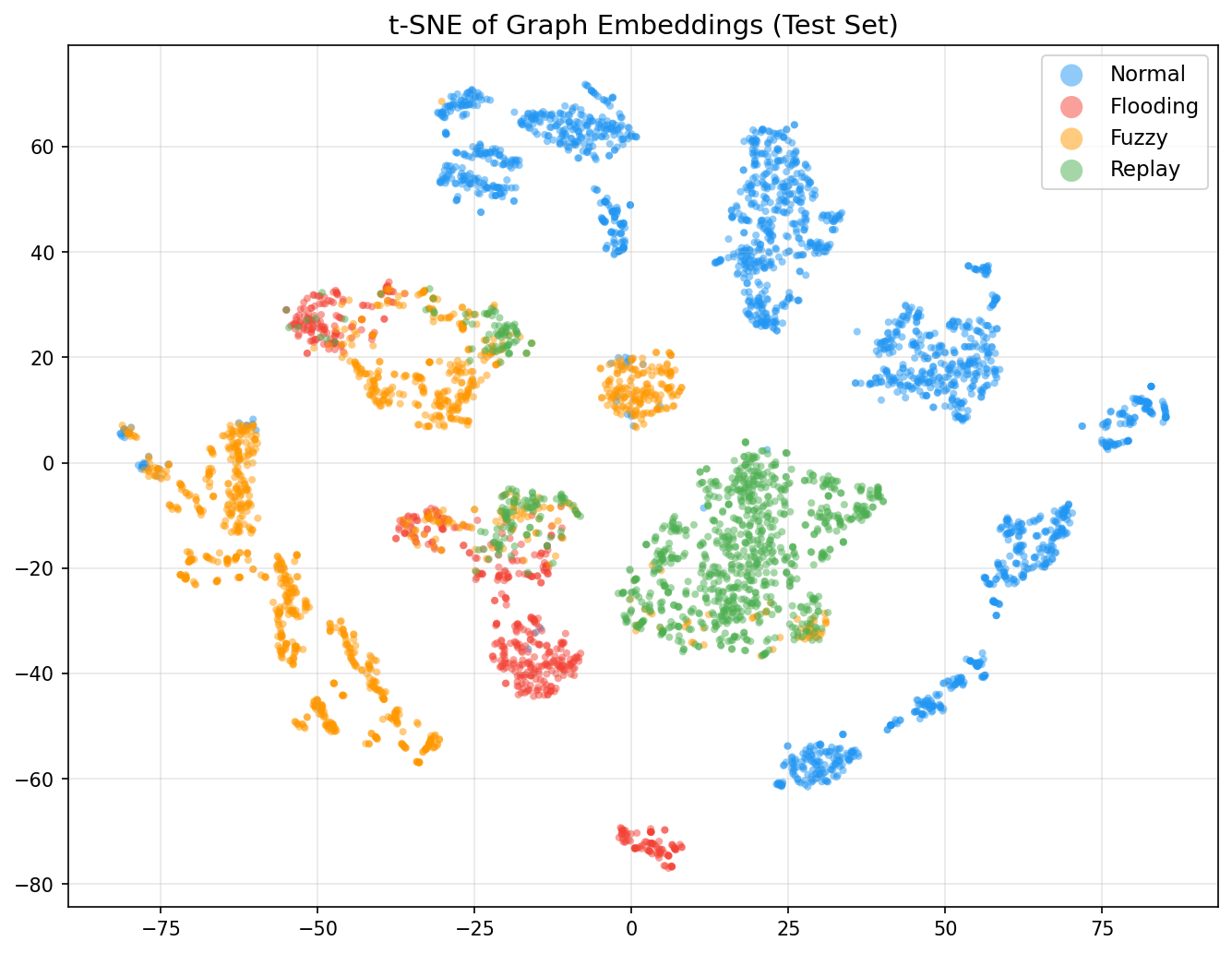}
\caption{t-SNE visualization of graph embeddings from the test set showing clustering by attack type.}
\label{fig:tsne}
\end{figure}

Fig.~\ref{fig:tsne} visualizes the learned graph embeddings projected to two dimensions using t-SNE. The four classes form well-separated clusters in the embedding space. Normal traffic (blue) occupies the upper and peripheral regions while attack types cluster in distinct zones. Some overlap between Flooding and Fuzzy clusters explains the confusion observed in the classification results. The clear spatial separation validates that the Residual GraphSAGE model learns meaningful graph-level representations that encode attack-specific structural patterns.

\subsection{Model Comparison}

Table~\ref{tab:comparison} compares the baseline GraphSAGE model (3-layer, 128-dim, CrossEntropy) with the proposed Residual GraphSAGE (4-layer, 192-dim, Focal Loss). The improved architecture with Focal Loss and augmentation achieves higher macro-F1 on validation and comparable overall accuracy while providing more balanced per-class performance.

\begin{table}[!t]
\centering
\caption{Baseline vs. Proposed Model Comparison}
\label{tab:comparison}
\begin{tabular}{@{}lcc@{}}
\toprule
Metric & Baseline & UAVCAN-GRAPH \\
\midrule
Architecture & GraphSAGE & Res-GraphSAGE \\
Hidden Dim & 128 & 192 \\
Layers & 3 & 4 \\
Loss Function & CrossEntropy & Focal ($\gamma$=2) \\
Augmentation & None & Fuzzy 2$\times$ \\
Val Macro-F1 & 0.9270 & 0.9327 \\
Test Macro-F1 & 0.9282 & 0.9329 \\
Test Accuracy & 0.9427 & 0.9425 \\
Binary F1 & 0.9937 & 0.9888 \\
\bottomrule
\end{tabular}
\end{table}

The marginal accuracy reduction alongside improved macro-F1 indicates that the proposed model sacrifices a small amount of majority-class performance to better detect minority attack types. This trade-off is desirable for security applications where missing a specific attack type carries greater risk than occasional false alarms on normal traffic.

\subsection{Discussion}

The experimental results demonstrate that temporal graph construction effectively captures the communication dynamics of CAN bus traffic under different attack conditions. Several observations merit discussion.

The UAVCAN-specific features (transfer ID continuity and toggle ratio) provide discriminative signals that complement traditional timing and payload features. Under normal operation the transfer ID increments sequentially and injected attack frames disrupt this sequence creating detectable anomalies in the graph node features.

Residual connections prove essential for the 4-layer architecture. Without skip connections the deeper model suffers from over-smoothing where node representations converge after repeated aggregation. The residual paths preserve input information and enable the model to learn both local and multi-hop structural patterns.

The choice of Focal Loss over standard cross-entropy addresses a fundamental challenge in attack detection: normal traffic vastly outnumbers each individual attack type. By down-weighting well-classified normal samples the model allocates more gradient signal toward the harder boundary cases between attack categories.

One limitation of the current approach is the fixed window size of 64 messages. Attacks with very slow injection rates may not produce sufficient anomalous packets within a single window to trigger detection. Adaptive window sizing or hierarchical multi-scale graph construction could address this limitation in future work.

\section{Conclusion}

This paper presented UAVCAN-GRAPH, a graph-based intrusion detection framework for multi-class attack classification on UAV CAN bus networks. By transforming temporal CAN traffic into directed weighted graphs and processing them through a Residual GraphSAGE architecture with Focal Loss optimization, the framework achieves strong detection performance across flooding, fuzzy and replay attack categories. The results on the HCRL UAVCAN Attack Dataset validate that graph-based representations combined with protocol-aware features provide an effective approach for UAV communication security. Future work will explore adaptive windowing strategies, attention-based edge weighting mechanisms and deployment optimization for resource-constrained UAV hardware.

\end{document}